# Deep Cellular Recurrent Network for Efficient Analysis of Time-Series Data with Spatial Information

L. Vidyaratne, *Member, IEEE*, M. Alam, A. Glandon, A. S. Shabalina, C. Tennant, and K. M. Iftekharuddin, Senior *Member, IEEE*

*Abstract*—Efficient processing of large-scale time series data is an intricate problem in machine learning. Conventional sensor signal processing pipelines with hand engineered feature extraction often involve huge computational cost with high dimensional data. Deep recurrent neural networks have shown promise in automated feature learning for improved time-series processing. However, generic deep recurrent models grow in scale and depth with increased complexity of the data. This is particularly challenging in presence of high dimensional data with temporal and spatial characteristics. Consequently, this work proposes a novel deep cellular recurrent neural network (DCRNN) architecture to efficiently process complex multi-dimensional time series data with spatial information. The cellular recurrent architecture in the proposed model allows for location-aware synchronous processing of time series data from spatially distributed sensor signal sources. Extensive trainable parameter sharing due to cellularity in the proposed architecture ensures efficiency in the use of recurrent processing units with high-dimensional inputs. This study also investigates the versatility of the proposed DCRNN model for classification of multi-class time series data from different application domains. Consequently, the proposed DCRNN architecture is evaluated using two time-series datasets: a multichannel scalp EEG dataset for seizure detection, and a machine fault detection dataset obtained in-house. The results suggest that the proposed architecture achieves state-of-the-art performance while utilizing substantially less trainable parameters when compared to comparable methods in the literature.

## I. Introduction

Typical pattern recognition applications oftentimes involve classification or regression of input data that is static in time. However, most real world data obtained through a set of observations almost always exhibit changes with time. Though in some cases, the change of observations in time can be ignored, certain applications that particularly deal with changes across time requires an additional temporal dimension to be incorporated in the pattern recognition process. Moreover, tasks such as monitoring multi-channel EEG for seizure detection and complex machine health monitoring may require recognition of patterns that extend in both spatial and temporal dimensions. Computational models that are specifically capable of capturing complex patterns in time and space are required to process such multi-dimensional time series data. One of the most challenging steps in constructing a machine learning model for complex time series analysis is an appropriate feature extraction scheme that effectively captures the patterns across time and spatial dimensions. These representative features may be a set of simple statistics of the time series data such as mean, variance, skewness, kurtosis, largest peak, and number of zero crossings. [1]. More descriptive features such as autoregressive coefficients [2], frequency power spectral features [3], and features derived from time-frequency analysis such as wavelet transform [3, 4], wavelet packet transform [4, 5], filter banks [6], and self-similarity features [7], and further engineered versions of these may also be considered to obtain a more discriminatory representation of data. However, one of the main problems associated with feature engineering is that the efficacy of such features essentially depend on the data, and the application. Therefore, the performance of a machine learning pipeline depends on the hand selection of a subset of features, or extraction of a set of new features based on the domain expertise. Feature learning with artificial neural networks (ANN) largely alleviates this problem by progressively learning the best possible discriminatory feature from data.

The availability of powerful computational tools and training methods have enabled deep neural networks to solve many difficult recognition problems in robotics [8], object recognition [9], and text recognition [10]. The typical feed-forward neural networks are predominantly used in processing data that is static in time due to its inability to process temporal relations owing to the limited forward information processing capability. Recurrent neural network (RNN) [11] or a time-

The authors would like to acknowledge partial funding of this work by the National Science Foundation (NSF) through a grant (Award# ECCS 1310353). Note the views and findings reported in this work completely belong to the authors and not the NSF.

The authors LV, MA, AG, and KMI are with the Vision Lab in Department of Electrical and Computer Engineering, Old Dominion University, Norfolk, VA 23529 (email:{lvidy001, malam001, aglan001, kiftekha} @odu.edu)
The authors AS, and CT are with the Jefferson Laboratory, Newport News, VA, 23606 (email:{shabalin, tenant}@jlab.org)

delay neural network (TDNN) is a variant of ANN with the added capability of information aggregation through feed-back connections. RNNs such as Elman and Jordan architectures [12-14] process time-series by reading samples sequentially in time, and the feed-back connections aid in retaining valuable information through time-steps. Further improvements to the feed-back units in retaining memory through longer time-sequences are tasked to Long Short-term Memory (LSTM) [15] units, and Gated Recurrent Units (GRU) [16]. Large-scale deep versions of recurrent neural networks have been successfully utilized in multiple domains [17-20]. Few works suggest the use of deep CNN and/or deep LSTM networks for processing EEG [21-23]. These typically involve an additional feature extraction step such as Fourier spectrum computation prior to the application of CNN for improved compatibility. The deep CNN is primarily used as a feature extractor while a LSTM layer is applied subsequently for temporal processing. However, the current state-of-the-art deep models suffer from a major limitation. The depth, complexity, and the amount of trainable parameters associated to these models grow proportionally to the complexity of the input dimensionality and the given task. This problem is further exacerbated in recurrent learning models as the additional feed-back links demands even more trainable parameters. Therefore, such architectures can grow prohibitively in the presence of large-scale, multi-source time-series data such as EEG.

Furthermore, the deep CNN and LSTM methods still largely ignore the spatial relevance in large scale time series data for some applications (e.g., EEG and machine fault detection) where space location information is of interest. The time series data recorded from different components in a machine health diagnosis, and fault detection system such as in [24] may hold spatial correlation based on the locality of the components. Specifically, the particle accelerator facility of the Jefferson Nation Labs contains multiple cavities situated serially on cryomodules [24]. Multiple RF signals from each cavity are recorded for monitoring the operating conditions [24]. Automated detection and classification of faults in this system involves efficient processing of time series data obtained from each cavity. In addition, EEG signal processing with conventional CNN and LSTM architectures faces similar challenges in literature. Interestingly, Thodoroff et. al. [22] proposes an image based representation combining Fourier spectral features from individual EEG electrodes into a single image based on the 2D projection of the EEG montage. This representation maintains the spatial locality of individual EEG electrodes to exploit the spatial relevance of seizure EEG. However, this is still processed using a large-scale multi-layer CNN and LSTM combined architecture that suffer from large computational cost for the networks.

Consequently, in order to address the general lack of computationally efficient methods for processing time series data that also maintain spatial relevance, this study proposes a novel deep learning architecture called deep cellular recurrent neural network (DCRNN). The DCRNN is inspired by the cellular neural network architectures [25-27] that is shown useful for real time image processing [26] and approximating the dynamic programming tasks [28]. The typical cellular architecture spans the area of a 2D input such as an image, overlapping each pixel with a corresponding cell in the network [27, 28]. Such cellular architectures enable distributed processing of information while maintaining synchronized communication with the neighboring cells. The cellular architecture also promotes extensive sharing of tunable parameters by placing identical neural structures in each cell [27]. We utilize this unique cellular sub-architecture in designing our DCRNN architecture for multi-dimensional time series data processing. The cellularity of the proposed architecture allows for processing sensor signals obtained from individual sources. The grid-like placement of cells in-turn enables communication with the neighboring cells, which allows learning spatial characteristics based on the locality of sensor signal sources. We also gain extensive trainable weight sharing by placing identical recurrent neural models within each cell. Moreover, the cellularity enables straightforward expansion of architecture for changes in the number of input sources, with only negligible increments to the number of trainable weights. We also investigate the versatility of the novel DCRNN architecture for classification of time series data from different application domains. Most typical deep learning architectures are purpose built to handle a particular task in a specific application domain. However, only a few studies that introduce new architectural developments [29, 30] in RNN show promise in the analysis of data from different application domains. Consequently we identify and leverage the particular similarities in the input multi-sensor data organization for these disparate domains to apply the proposed DCRNN. Accordingly, we introduce a novel cellular recurrent neural network architecture, and investigate the performance of the proposed architecture using two representative datasets from two different application domains: 1) a multi-channel scalp EEG dataset for seizure detection, 2) an in-house dataset acquired from Jefferson National Lab on machine fault detection.

Section II of this paper provides the background information on cellular neural network architecture and long short-term memory units. Section III introduces the novel deep cellular recurrent neural network architecture and the training paradigm with input dataset preparation procedure. Section IV presents and analyzes the results with comparison to state-of-the-art methods in literature. Finally, section V provides the conclusion and future work.

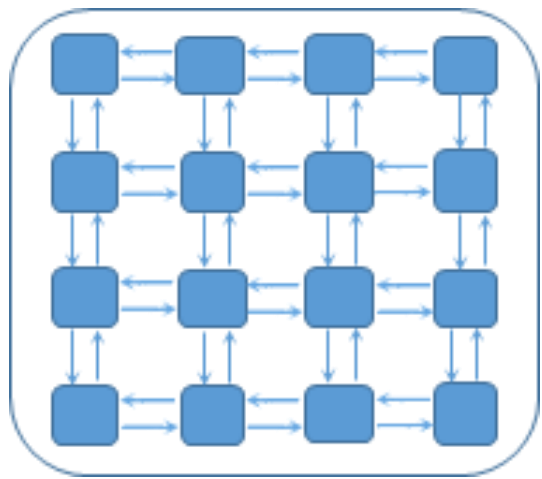

Fig. 1. Cellular neural network architecture. This cellular neural network consist of 16 cells in a 4 × 4 2D grid arrangement.

## II. BACKGROUND REVIEW

### A. Cellular Neural Network

Cellular neural network is an architecture that consists of multiple cells with elements arranged in a geometric pattern [25, 28]. Each element in the cellular neural network may house a single neuron or a complex ANN. However, these elements are usually made with identical sub-structure to maximize the shareability of trainable weights among the cells. A typical cellular network architecture spanning a 2D space is shown in Fig. 1.

The architecture shown in Fig. 1 can be used to process an input that consists of sensor signal sources in a 2D spatial arrangement with size $4 \times 4$. In this arrangement, each cell may process the individual inputs of the corresponding sensor signal source. Additionally, as shown in Fig.1, each cell in the cellular architecture consists of communication channels between the neighboring cells. These channels may allow for processing the local geometric patterns exhibited among sensor signal sources within multi-dimensional time series data.

### B. Long Short-term Memory Networks

The generic recurrent neural networks are known to suffer from limited reach of context over time series data in generating the network output. This is due to the limited or decaying backpropagation error over long time periods of a given time series [31]. This can be considered as a vanishing gradient problem over time, similar to the vanishing gradient problem that occurs over depth of a deep network architecture. Consequently, the Long short-term memory is developed to address this vanishing error signal, with the introduction of memory gates that control the flow of context over time [32]. Figure 2 shows a signal flow diagram of an LSTM unit.

The full operation of an LSTM unit for a single time step [33] is described in Eqns. (1) to (5).

$$i_t = \sigma(W_i x_t + U_i h_{t-1}), \quad (1)$$

$$f_t = \sigma(W_f x_t + U_f h_{t-1}), \quad (2)$$

$$o_t = \sigma(W_o x_t + U_o h_{t-1}), \quad (3)$$

$$s_t = f_t \odot s_{t-1} + i_t \tanh(W_s x_t + U_s h_{t-1}), \quad (4)$$

$$h_t = o_t \odot \tanh(s_t); \quad (5)$$

Typical inputs for an LSTM at time step $t$ includes the signal input $x_t$, hidden output of the previous time step $h_{t-1}$, and memory accumulated at the previous time step $s_{t-1}$. The input signal $x_t$ and previous hidden signal $h_{t-1}$ are combined in Eqns.

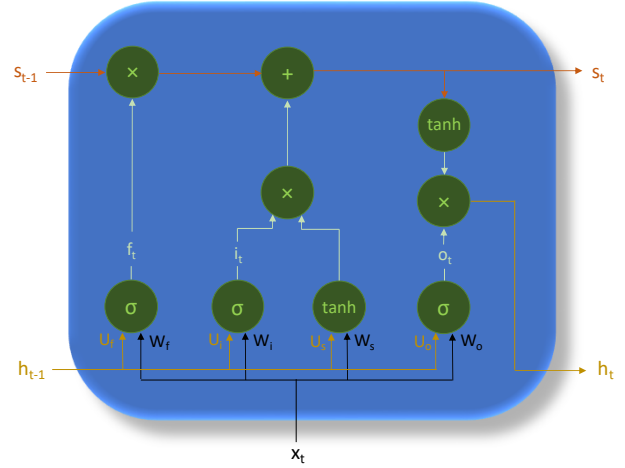

Fig. 2. Long Short-Term Memory Unit Signal Flow Diagram

(1) to (3) and passed through a sigmoid activation function to obtain $i_t, f_t$ and $o_t$. These are known as the "gates" such that if the sigmoid output is near 0, the gate signals have the effect of inhibiting the propagation of the corresponding input signal. Accordingly, the input gate $i_t$ is used to control the effect of the signal input. The forget gate $f_t$ is used to clear the memory. The output gate $o_t$ is used to clear the hidden output. The effect of the three gates $i_t, f_t$ and $o_t$ on the running memory $s_t$, and the hidden output $h_t$ can be observed in Eqns. (4) and (5). This gate combination in LSTM helps preserve the long term and short term temporal relevance in time sequences of variable length [32].

While the LSTM is able to build contextual memory through time, this context at time step $t$ is limited to at most from time step 0 to the current time step $t$ and the generic LSTM do not make use of the future context (such as $t + 1 \ to \ T$) in processing $x_t$. The bidirectional LSTM (BLSTM) [31, 34] alleviates this problem by utilizing the past and future context when the whole time sequence is available. The BLSTM is an extension to the generic LSTM where two different LSTMs process the time series from forward ($LSTM^{d1}$) and backward ($LSTM^{d2}$) directions respectively. The BLSTM then combines the outputs from each using an additional layer to obtain the final output [31].

## III. METHODOLOGY

This section describes the proposed DCRNN architecture, procedure.

### A. The proposed DCRNN Architecture

The proposed DCRNN architecture is shown in Fig. 3. Note that the cellular front end of the proposed architecture is expanded to overlap a multi-source 2D input pattern as illustrated in Fig. 3. This enables the LSTM network core in each cell to process the time series data generated from the corresponding sensor signal simultaneously. The LSTM core network within each cell can be configured as needed for a particular task. However, we constrain the LSTM core

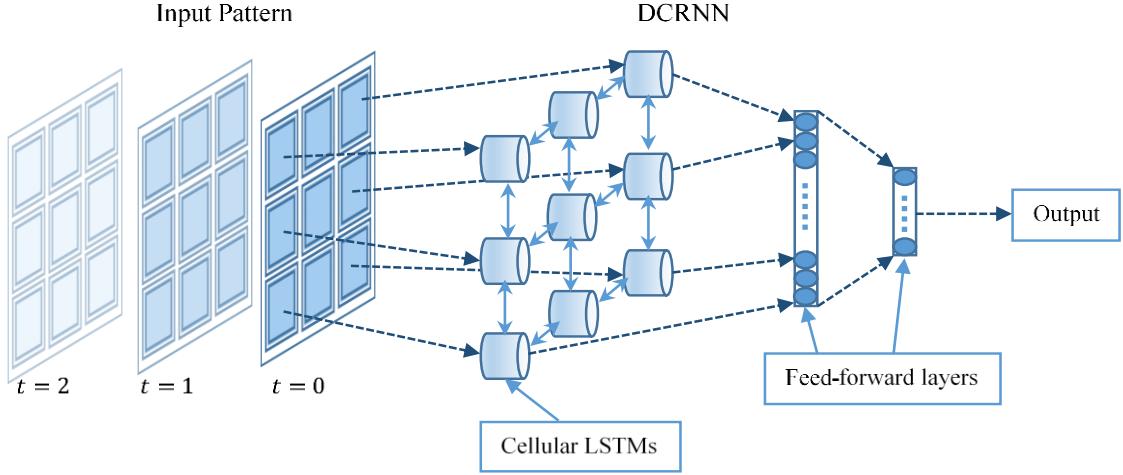

Fig. 3. Proposed DCRNN architecture. Each cell in the cellular sub-architecture hold a configurable LSTM network. Final outputs of each cell is aggregated and passed through a feed-forward network followed by classification.

architecture to be identical for each cell to ensure maximum trainable weight sharing. This novel DCRNN model, therefore, offers versatility of cellular neural processing combined with flexible time series processing of recurrent LSTM while keeping the spatial location information of input sensor signal. the associated spatial organization of input data and training

It is also evident from Fig. 3 that communication paths exist between a given cell and its four neighboring cells. The neighborhood information processing occurs at each time step. For instance, consider the cell $j,k$ of the cellular grid of size $J \times K$ is processing a time series at time step $t$. Along with the input of time series at $t$, we configure an additional path to the core architecture coming from the neighbors $((j-1,k),(j+1,k),(j,K-1),(j,k+1))$ outputs obtained at time $t-1$. In order to accommodate this additional neighbor information path in a 2D cellular setting, we augment the LSTM equations taking the core at cell $j,k$ as follows:

$$i_{j,k,t} = \sigma(W_i x_{j,k,t} + W_{Ni} N_{j,k,t} + U_i h_{t-1}), \quad (6)$$

$$f_{j,k,t} = \sigma(W_j x_{j,k,t} + W_{Nf} N_{j,k,t} + U_f h_{t-1}), \quad (7)$$

$$O_{j,k,t} = \sigma(W_o x_{j,k,t} + W_{No} N_{j,k,t} + U_o h_{t-1}), \quad (8)$$

$$S_{j,k,t} = f_{j,k,t} \odot S_{j,k,t-1}$$
$$\odot i_{j,k,t} tanh(W_s x_{j,k,t} + W_{Ns} N_{j,k,t} + U_s h_{t-1}), \quad (9)$$

$$h_{j,k,t} = O_{j,k,t} \odot \tanh(S_{j,k,t}). \quad (10)$$

Here,
$$N_{j,k,t} = [h_{j-1,k,t-1}, h_{j+1,k,t-1}, h_{j,k-1,t-1}, h_{j,k+1,t-1}]. \quad (11)$$

Note that the previous time-step hidden output information of the four closest neighbors given in eq. (11) are used as an additional input signal $N_{j,k,t}$ for the LSTM network at each cell. With a $G \times 1$ dimensional hidden output per cell, we typically assign just one neuron output (the $G^{th}$ element) as the output for neighbors. Though this is configurable, we have experimentally determined that a single neighbor output per cell is sufficient for adequate performance.

**Algorithm 1:** Training procedure of DCRNN

**Initialization:**
- Set cellular sub-structure parameters $W_c \in \{W_i, W_j, W_o, W_s, W_N\}$ and feed-forward parameters $W_{ff}, W_{\bar{y}}$ with random values

**Training:**

for each epoch
  for each sample/batch
    for each time step $t \leftarrow 0$ to $T$
      1. propagate through the cellular sub-net:
      for each cell $j,k \leftarrow 0$ to $J$ and $0$ to $K$:
- Obtain corresponding signal input $x_{j,k,t}$ and input from neighbors $N_{j,k,t}$
- Compute LSTM memory $S_{j,k,t}$ and hidden output $h_{j,k,t}$ using Eq.(6)-(10)

      end
    end
    2. Propagate through feed-forward sub-net:
- Aggregate $h_{j,k,T}$ from each cell $j,k$ to form $H$
- Compute output $\bar{y}$ using Eq.(12),(13)
- Compute $E$ using Eq. (14)

    3. Perform DCRNN back-propagation:
- Use standard back propagation to obtain $\Delta W_{\bar{y}}$, and $\Delta W_{ff}$
- Use BPTT to obtain $\Delta W_c$:
      for each time step $t \leftarrow T$ to $0$
        for each cell $j,k \leftarrow J$ to $0$ and $K$ to $0$
          $\Delta W_c = \Delta W_c + \nabla_{W_c} E$
      end
    end
    4. Update the Weight parameters:
- $W_{\bar{y}}(new) = W_{\bar{y}}(old) - \alpha * \Delta W_{\bar{y}}$
- $W_{ff}(new) = W_{ff}(old) - \alpha * \Delta W_{ff}$
- $W_c(new) = W_c(old) - \alpha * \Delta W_c$
  end
end

The cellular configuration makes it necessary to hold cell specific intermediate, final hidden and memory outputs as shown in Eqns. (6) to (10). However, maintaining identical LSTM settings for each cell allows sharing of trainable parameters. Though only shown for a single LSTM layer, the cell core architecture can be expanded for multiple layers or bidirectional processing as necessary. The final outputs at time step $T$ of each cell $h_{j,k,T}$ are aggregated to obtain the feature vector $H$. Subsequently, the feature vector $H$ is passed through the feed-forward sub-net to obtain the final output as follows:

$$FF = \sigma(W_{ff}H + b_{ff}), \quad (12)$$
$$\bar{y} = softmax(W_{\bar{y}}FF + b_{\bar{y}}), \quad (13)$$

Given the ground truth classification as $y$, the classification error $E$ is computed using the Mean Squared Error based loss-function:

$$E = \frac{1}{2}\|y - \bar{y}\|_2^2; \quad (14)$$

The training of the network is performed by obtaining partial derivatives of feed-forward weights $\Delta W_{\bar{y}}$ and $\Delta W_{ff}$ using standard back-propagation algorithm, and $\Delta W_c$ using back-propagation through time across all cells. The detailed training procedure of the proposed DCRNN architecture is shown in Algorithm 1.

### B. Computational complexity Analysis for DCRNN

One clear advantage for DCRNN is the extensive use of weight sharing in the cellular recurrent sub-architecture as shown in Fig. 3. This is evident especially when the DCRNN is used to process time series data with multiple sensor signal sources spread in 2D space. Consider a time series data sample at time-step $t$ with $J \times K$ individual signal sources spread in a 2D space. The total number of parameters ($N_{DCRNN}$) of the DCRNN architecture is given by,

$$N_{DCRNN} = (n_{CLSTM} \times m) + (J \times K \times n_{ff}) + c \times n_{ff} \quad (15)$$
(LSTM weights in a cell) (feed-forward weights) (classifier)

Whereas, the required number of parameters ($N_{DLSTM}$) of a deep LSTM with similar depth is given by,

$$N_{DLSTM} = (n_{LSTM} \times m \times J \times K) + (n_{LSTM} \times n_{ff}) + c \times n_{ff} \quad (16)$$

Considering the LSTM network contains multiple trainable weights as shown in Eq. (1) to (5), the upper bound of the required number of parameters for the generic deep LSTM (DLSTM) in presence of above data is $O(n_{LSTM} \times m \times J \times K)$ where $m$ denotes the dimensionality of the data in a single signal source. Conversely, the cellular architecture with weight sharing manages to process the same data with just $O(n_{CLSTM} \times m)$ complexity. Further note that typically $n_{LSTM} \gg n_{CLSTM}$ due to the large sensor signal input dimensionality faced by the generic DLSTM architecture. In contrast, the DCRNN requires very small amount of recurrent LSTM core units within each cell as the cellular architecture processes data from each sensor signal source separately.

### C. DCRNN Based Multi-channel EEG Processing Pipeline

As discussed in Section I, multi-channel scalp EEG data exhibits the characteristic of time series with spatial locality. The spatial locality may specifically be of interest in automated EEG signal processing as EEG signal collected at different locations in brain represents specific seizure activity [22]. Accordingly, we utilize a multi-channel scalp EEG dataset known as the CHB-MIT EEG database [35]. This dataset consists of long-term multi-channel EEG recorded from multiple pediatric patients with intractable seizures. More importantly, the scalp EEG setup used in most cases contain 23 bipolar EEG signals recorded from individual electrodes placed

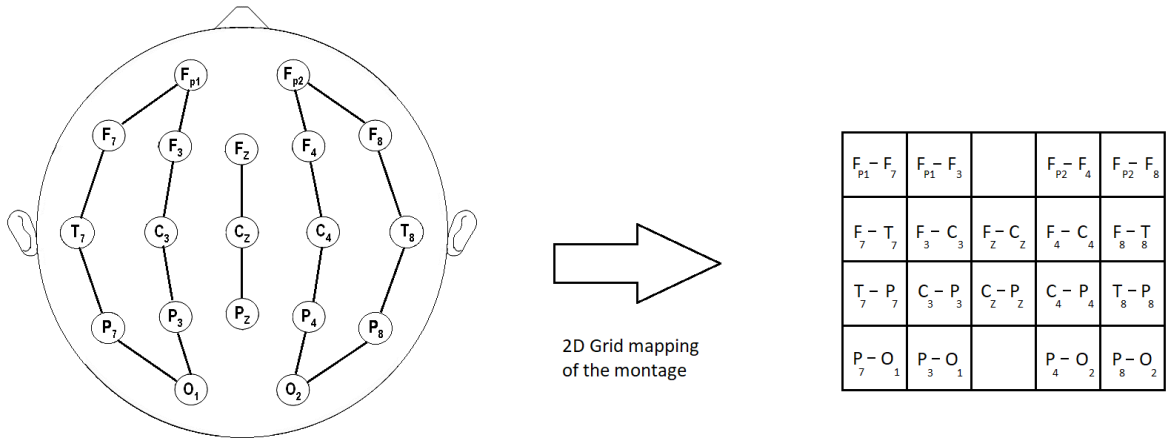

Fig. 4. The 2D grid mapping of the long-term bipolar EEG montage used in the CHB-MIT scalp EEG Dataset. The 2D grid approximation is a typical input to the proposed DCRNN.

according to the International Federation of Clinical Neurophysiology 10-20 system.

For effective processing of the EEG with spatial orientations intact, we map the EEG montage with 18 representative bipolar channels into a 2D grid setting for better visualization as shown in Fig. 4. Note that the raw EEG signals localized as shown in Fig. 4 matches with a 2D spatial input arrangement required for the proposed DCRNN architecture in an input grid arrangement of size $J = 4$ and $K = 5$. Note the mapping in Fig. 4 is scalable that any additional signal sources (channels) may be easily accommodated by rearranging the specified grid. This simply expands the cellular arrangement of the DCRNN correspondingly without additional complexity due to weight sharing. We utilize this dataset arrangement with the proposed DCRNN architecture to perform automated seizure detection.

*D. DCRNN Based Machine Fault Detection Pipeline*

In order to investigate the versatility of the proposed DCRNN architecture, we utilize a second dataset for machine fault detection. The dataset is derived from a database maintained by the Jefferson National Laboratory based on the hardware specific faults encountered in the particle accelerator facility. A brief description of the hardware arrangement is as follows. The Continuous Electron Beam Accelerator Facility (CEBAF) at Jefferson Laboratory incorporates multiple cryomodules with superconducting radio frequency (SRF) cavities. Each cryomodule contains eight such cavities connected serially. A fault that occurs in any of these cavities disrupt the experimentation at the CBAF facility. More information on the facility, the hardware, and the associated data can be found in [24]. In summary, multiple radio frequency (RF) signals are recorded from each SRF cavity in each cryomodule and a database of recording with cavity faults are maintained for further study.

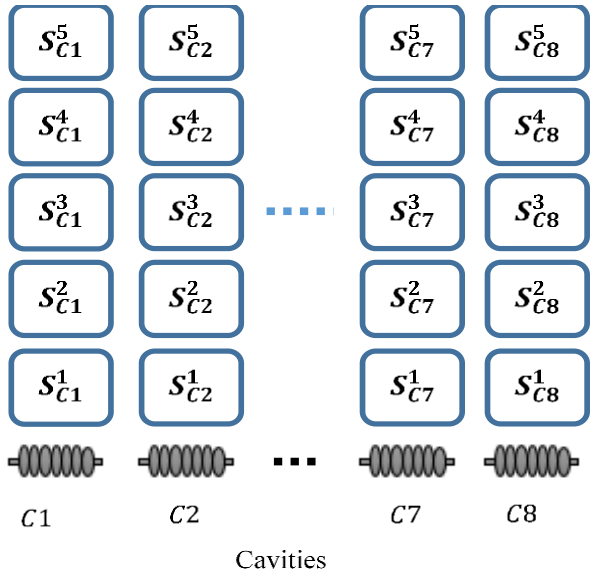

Fig. 5. The 2D time series grid arrangement of the Jefferson Lab cavity fault detection database.

We utilize this database for automated multi-class fault detection with the proposed DCRNN. The cavities are arranged in a serial fashion within the cryomodule. We select 5 representative RF time series signals per cavity based on expert recommendation. We subsequently map the eight cavities and corresponding RF signals in a 2D grid layout as shown in Fig. 5. With this mapping, the 5 time series data from each cavity is separated in rows while the serial cavity arrangement is preserved in columns. This ultimately obtains a grid of size $J = 5$ and $K = 8$, and an efficient 2D arrangement for the proposed DCRNN architecture.

*E. Network Preparation*

This section summarizes the input specific configuration of the proposed DCRNN architecture.

First, the DCRNN architecture is prepared for analysis with EEG dataset as follows. We first implement the cellular recurrent architecture based on the EEG input mapping shown in Fig. 4. In the cellular sub-net, we implement a bidirectional LSTM architecture with 5 LSTM units in each direction. Note that the bidirectional LSTM architecture is made identical in all cells to allow sharing of trainable weights. The outputs from bidirectional architecture is aggregated across all cells and passed to the first feed-forward layer consisting of 50 neurons. The final classification layer configured for two class classification (seizure vs. non-seizure EEG) with softmax activation. The other feed-forward layers utilize sigmoid activation as discussed in Eq. (12). With this setup, each 1 second segment of EEG is classified as either normal or seizure EEG.

Subsequently, the DCRNN architecture is reconfigured for the machine fault detection data analysis as follows. We implement the cellular recurrent architecture to complement the data mapping arrangement in Fig. 5. Accordingly, the cellular sub-architecture contains 40 individual cells in $5 \times 8$ configuration. Within each cell, we setup a unidirectional LSTM architecture consisting of 5 LSTM units. Similar to EEG, the LSTM sub-sub-architecture is made identical in each cell to ensure full weight sharing. Final outputs of LSTMs in each cell is aggregated and processed through a feed-forward layer consisting 100 neurons following Eq. (12). The final classification layer is configured for a 5 class classification task with softmax activation. We classify each of the ~600 waveform events based on the corresponding fault class.

## IV. RESULTS AND DISCUSSION

We evaluate the performance of the proposed DCRNN model using two multi-dimensional time series datasets: CHB-MIT scalp EEG [35], and the Jefferson Lab machine fault dataset [24] as discussed above. We specifically select these datasets from different application domains to evaluate the scalability of the proposed DCRNN architecture in multiple application domains.

## A. Scalp EEG Dataset

As introduced in Section III C, the CHB-MIT scalp EEG dataset consists of long-term bipolar referenced multi-channel EEG recorded from pediatric patients with epileptic seizures. We utilize EEG data from 20 patients containing 124 separate seizure events for the analysis. The EEG is recorded in continuous segments of 1 to 4 hour duration. All EEG time series signals are sampled at 256 Hz. The Seizure events within the long-term EEG segments are annotated by an expert [35]. We perform patient specific seizure detection using the proposed DCRNN model.

The EEG preparation for analysis is as follows. We extract and segment all available raw seizure EEG into 1 second segments. We subsequently segment the non-seizure EEG into 1 second segment and perform randomized under sampling to obtain a patient specific dataset of seizure and non-seizure EEG. The dataset is then prepared with the mapping procedure specified in section III C for analysis. Note that we simply normalize the raw EEG without any additional pre-processing or feature extraction for this analysis. The patient specific dataset is finally utilized in a 5-fold cross validation procedure to obtain the performance of the proposed architecture.

Figure 6 summarizes the patient specific EEG classification results obtained with the DCRNN architecture. According to Fig. 6, seizure detection accuracy for most patients are well over 90%. Specifically, the DCRNN achieves an average accuracy of 91.3% with a median of 92.1%. However, when seizure detection criterion is considered, sensitivity score plays a more important role. This is due to the fact that in a realistic setting, one would expect to correctly identify all seizure events even at the cost of a relatively higher false positive numbers. Consequently, the proposed architecture achieves an average sensitivity value of 94% with a median sensitivity of 95%. The DCRNN model still manages to maintain a median specificity value of 90.5%. The proposed model also achieves a mean and median F1 scores of 91.4% and 92.25% respectively.

Table I compares the seizure detection performance of the proposed DCRNN model with other studies in literature. Table I shows that the proposed architecture manages to achieve comparable seizure detection performance to other state-of-the-art methods in literature. Specifically, the sensitivity of seizure detection is only slightly lower than the methods in [38] and [5], and higher than that of [40], all of which utilizes the same dataset. However, we point out that these methods use several complex feature extraction, feature selection, and classification methods in the associated seizure detection pipelines. For example, the seizure detection algorithm in [5] uses fractal dimension and harmonic wavelet packet transform on each EEG signal to extract features, and subsequently utilizes relevance vector machine (RVM) classifier for seizure classification. The method in [38] constructs a filter bank to extract features from each EEG signal followed by classification with SVM. The method in [40] uses several features based power spectral density (PSD) measures and temporal statistics of EEG time series data followed by feature selection method such as principal component analysis (PCA) and linear discriminant analysis (LDA). The study then experiments with several classification models such as linear discriminant classifier, quadratic discriminant classifier, polynomial classifier, etc., and obtains the best sensitivity of 93% with a K-nearest neighbor (KNN) classifier. Yao et. al. [41] proposes a deep recurrent learning approach for seizure detection using the same dataset. The architecture includes 15 recurrent layers with different time-scale hierarchies that are composed of 128, 200, and 250 hidden recurrent units for each 5 layer block, respectively. Similarly, Park et al. [42] utilizes a 7 layer CNN architecture containing both 1D and 2D convolutional filters to process a multi-channel EEG dataset for seizure detection.

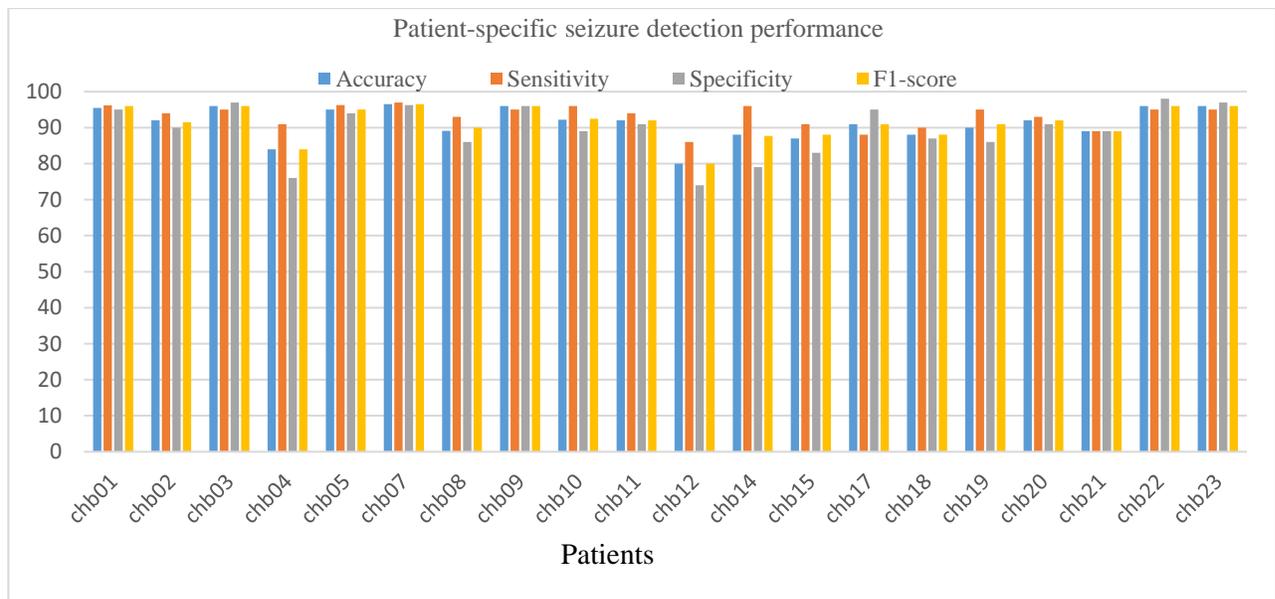

Fig. 6. Summary of patient specific seizure detection performance of the proposed DCRNN model

TABLE I
PERFORMANCE COMPARISON OF THE PROPOSED DCRNN MODEL WITH OTHER METHODS ON SEIZURE DETECTION WITH SCALP EEG

| Methods | Sensitivity (%) | Specificity (%) | Accuracy (%) | Conventional Feature Engineering and Classification | Automated Feature Learning and Classification |
|---|---|---|---|---|---|
| Yuan et. al. [36] | 93.8 | 94.8 | 94.9 | Yes | No |
| Subasi et. al. [37] | 93.10 | 92.8 | 93.1 | No | Yes |
| Shoeb et. al. [38] | 96 | - | - | Yes | No |
| Vidyaratne et. al. [5] | 96 | - | - | Yes | No |
| Khan et. al. [39] | 83.6 | 100 | 91.8 | Yes | No |
| Fergus et. al. [40] | 93 | 94 | - | Yes | No |
| Yao et. al. [41] | 87.3 | 86.7 | 87 | No | Yes |
| Park et. al. [42] | 80.8 | 91.7 | 85.6 | No | Yes |
| **DCRNN** | **94** | **90** | **91.3** | **No** | **Yes** |

TABLE II
PERFORMANCE COMPARISON OF THE PROPOSED DCRNN MODEL WITH OTHER METHODS ON MACHINE FAULT DETECTION DATASET

| Methods | Number of recurrent units in each layer | 10-fold accuracy ± standard deviation |
|---|---|---|
| AR features + SVM | - | 90% ± 4% |
| AR features + RF | - | 91.5% ± 2% |
| AR features + LR | - | 87.4% ± 4.8% |
| Deep LSTM | 256-256 | 88.83% ± 2.4% |
| **DCRNN** | **5-5** | **89.1% ± 2.7%** |

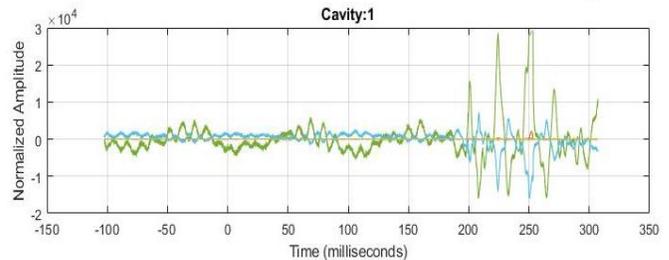

Fig. 7. Example RF waveforms extracted from cavity 1.

In contrast, the proposed DCRNN contains only 5 bidirectional LSTM units in the recurrent hidden layers of each cell. With cellular weight sharing, we maintain the number of units the same among all cells that process corresponding channels. This comparison shows the highly superior computational efficiency of the proposed architecture. In summary, the proposed architecture performs efficient feature learning and classification simply utilizing minimally pre-processed EEG. Moreover, time series processing with LSTM is performed within the cellular sub-net, which allows for simultaneous processing of each EEG channel while taking into account the locality of electrodes on the scalp. Minimal pre-processing with automatic feature learning and efficient use of trainable weights make DCRNN desirable for multi-channel EEG processing applications.

B. *Machine Fault Detection Dataset*

The Jefferson Labs machine fault detection dataset includes approximately 600 samples of cavity waveform data acquired from the particle accelerator system. Each sample contains 17 RF waveforms recorded from each of the 8 SRF cavities. Each waveform contains ~1.6 seconds (8196 time samples) of data that includes system failure due to a certain fault event. The dataset is inspected and categorized into 5 known fault types by an expert. An example waveform extracted from cavity 1 is shown in Fig. 7.

We prepare the machine fault data for analysis as follows. We select 5 most significant RF waveforms for analysis based on visual analysis by an expert. We subsequently normalize the waveforms based on the z-score normalization technique. Even though the RF waveforms are sampled at a very high rate, we observe that the actual fault event is a relatively low frequency event. Therefore we perform aggressive down sampling of the selected waveforms by a factor of 20 to obtain time series data of approx. 410 time samples. The data is subsequently arranged based on the mapping introduced in section III D and visualized in Fig. 5. The dataset is utilized in a 10-fold cross validation process to obtain the performance of the proposed DCRNN architecture.

In order to compare the performance of DCRNN on the fault classification dataset, we construct bidirectional LSTM architecture with two 256 LSTM units each followed by a feed forward layer of 512 neurons and a 5 class classification layer. Additionally, we set up a machine learning framework for the fault classification task. For this, we perform feature extraction on 5 selected waveforms utilizing autoregressive (AR) analysis. Accordingly, we obtain a 6-dimensional feature vector per waveform to construct a 240 ($6\ features\ \times 5\ waveforms\ \times 8 cavities$) element feature vector for each data sample. We subsequently perform 10-fold cross validation analysis using classifiers such as Logistic regression (LR), support vector machine (SVM), and Random Forrest (RF). The 10-fold cross validation performance of the proposed architecture performance of the proposed along with comparison with other methods are shown in Table II.

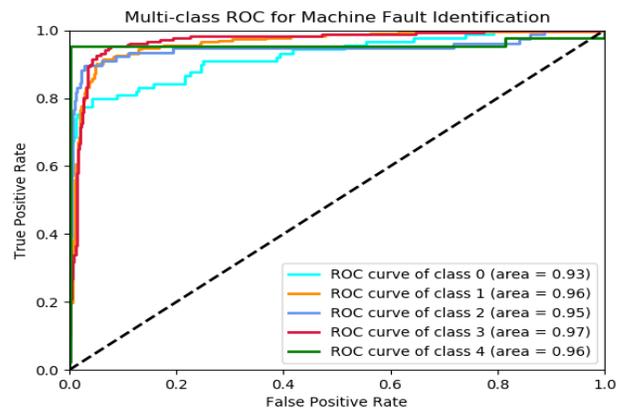

Fig. 8. ROC performance curve of DCRNN for multi-class machine fault detection.

As shown in Table II, between the two deep learning models, the proposed DCRNN offers comparable accuracy. However, note the large difference in hidden LSTM units used for the recurrent layers in both deep LSTM and DCRNN. This is due to the cellular processing feature that maintains the location information for sensor signal in DCRNN as illustrated in Fig.3. Therefore, the input dimensionality of the sensor signal per cell is quite small, and will require much smaller number of LSTM units per cell. Moreover, since the LSTM architecture is shared among cells, the number of trainable parameters does not grow in size. The ROC curve of DCRNN for multi-class processing is shown in Fig. 8. The area under the curve is consistently near unity for all 5 classes indicating that our algorithm provides high sensitivity and specificity, without a need to sacrifice either.

Though the machine learning based methods in Table II perform slightly better than that of the proposed DCRNN model, we point out that the associated pipeline requires autoregressive feature extraction from each RF waveform of each cavity. This may be a tedious and computationally intensive process, especially if the number of waveforms or the cavities are higher. The proposed DCRNN architecture is quite helpful in this regard as it simply requires to expand the cellular grid to accommodate the increased input sources. Finally, the trainable weight sharing property of the cellular architecture in the proposed model helps to minimize the computational complexity as analyzed in section III B.

## V. Conclusion

This work proposes a novel deep cellular recurrent neural network (DCRNN) architecture for efficient processing of large-scale time-series data with spatial relevance. The DCRNN model consists of a cellular recurrent core-network that operates in 2D space to enable efficient processing of time series data while considering signals from multiple spatially distributed signal sensors sources. The cellular architecture processes data from each localized sensor signal source individually in a synchronized manner. This 2D distributed processing approach enables minimum use of recurrent LSTM units within each cell due to the locally reduced input dimensionality. Moreover, time series data obtained from spatially distributed sensor systems such as multi-channel EEG may hold importance in the locality of the sensor signal for many associated tasks. The cellular architecture of the proposed DCRNN preserves the locality of the distributed sensor signals by mapping itself onto the 2D space. The inter-cellular weight sharing property further improves the efficiency of the proposed model. We further investigate the adaptability and the versatility of the proposed architecture in processing time series data from different application domains. Accordingly, the performance of the proposed DCRNN model is evaluated using two large-scale time series datasets obtained from biomedical, and machine fault analysis application domains, respectively. The results show that the proposed architecture achieves state-of-the-art performance with respect to comparable machine learning and deep learning methods in both domains while utilizing significantly less amount of recurrent processing units and trainable parameters.

Our future plan is to further improve the data processing efficiency of the proposed architecture by introducing parallelized processing within the cellular architecture. Such improvements in the implementation may allow for the proposed architecture to be deployed for real-time processing with resource restricted embedded hardware systems.

## Acknowledgement

The machine fault detection data for this study come from Jefferson Science Associates, LLC under U.S. DOE Contract No. DE-AC05-06OR23177